# WEIGHTED ACCURACY ALGORITHMIC APPROACH IN COUNTERACTING FAKE NEWS AND DISINFORMATION

**Kwadwo Osei Bonsu, M. Eng**

*Zhejiang Gongshang University,*

*School of Economics,*

*School of Law and Intellectual Property*

*k.oseibonsu@pop.zjgsu.edu.cn*

## Abstract

As the world is becoming more dependent on the internet for information exchange, some overzealous journalists, hackers, bloggers, individuals and organizations tend to abuse the gift of free information environment by polluting it with fake news, disinformation and pretentious content for their own agenda. Hence, there is the need to address the issue of fake news and disinformation with utmost seriousness. This paper proposes a methodology for fake news detection and reporting through a constraint mechanism that utilizes the combined weighted accuracies of seven machine learning algorithms.

## 1-INTRODUCTION

In this era of digital communication, it is very easy to get noticed from the public by addressing public grievances and demands regardless of the authenticity to the claim. This type of content can be termed as fake news. Fake news is written and published in order to gain readers' attention. This usually misleads readers by financially or politically sensationalist or exaggerated false headlines (Wiki 2020).

Most fake news detection systems try to predict the probability that a particular news report, editorial or some online content is intentionally produced for deceptive purposes (Chen; Conroy and Rubin 2015); for example, the fake news of Steve Jobs' heart attack had worst impact on the stock prices of Apple Inc. The detection mechanisms try to find and analyse linguistic clues based on the assertion that the language of truth is different from the language of lies (Feng and Hirst 2013) (Bachenko; Larcker and Zakolyukina 2012) and that liars are more psychologically prone to use emotionally driven sentences but less self-oriented pronouns.

One of the biggest reasons why fake news has thrived in society is that humans fall victim to Truth bias, Naïve Realism and Conformation bias. The major classes and goals, along with the motive of proposing an approach to the detection system design have emerged by utilizing disparate techniques. Currently, there are two major categories of method approaches which are Linguistic and Network Approaches.

## 2-LITERATURE REVIEW

Many experts have expressed that fake news is not just exclusive to marketing and public relations but also greatly associated with political propaganda and cultural engineering (ASIST 2015). Traditionally, marketing, advertisement and public relations are the major departments that have been dealing with the implications arising from fake news and disinformation. Some experts have also suggested that a competitive corporate environment could be a solution for reducing the influx of fake news in the information environment (Balmas 2014).

The inception of the world wide web and advancement in information technology has also groomed a breed of perpetrators of fake news who are motivated by attracting web traffic that has been a disastrous recipe for inflicting woes on web users. Some of the content published on these sites are intentionally coded with malware and viruses that are hidden. Companies especially have been a milky target for cyber-attacks (ICFNDS 2018). Apart from the potential threat from malwares and viruses, many of the modern organizations are also concerned with employee manipulation for giving out their credentials. There is the lobby of content publishers that are using click baits to facilitate their phishing objectives (Nah 2015).

Database and data integrity are the information technology security implications associated with the practices of fake news (Pogue 2017). Data is a valuable asset in the digital era of information technology and artificial intelligence and the protection and preservation of data is of utmost importance. In some cases, fake news has been beneficial to some companies where there were reports of positive hike among various stocks of those companies as a result of fake news (Hassid 2011).

Fake news may help to enhance the marketing objective of an enterprise when the information is in the interest of the company, though in reality there might be no such extraordinary services available from these companies and their reputation could skyrocket if it goes in their favor. There could however be ramifications if the propaganda is exposed, such firms could be bound to dysfunction in the long-time frame.

Finally, there are ethical concerns regarding the whole concept of fake news propagation (which will not be further discussed in this paper). Journalists and writers are generally supposed to furnish concrete information, news and reports to the world by organizing authentic raw data materials. They are responsible for providing accurate information, this is why it is alarming that for the sake of popularity many of them ignore their code of conduct.

This paper seeks to propose a new methodology for infusing the weighted performances of different machine learning algorithms in detecting and reporting fakes news and misinformation.

**3-EXISTING FAKE NEWS DETECTION MECHANISMS**

There are various methods proposed and/or applied for fake news detection. Some of them are discussed below:

a) Network mode method is a working principle of network analysis. What makes this mode different from other methods is its requirement of extensive data collection that assesses truth from new sentences (Conroy; Rubin and Chen 2015). This method checks the claims being made in the news or articles and their status of being real or fake determined with respect to authenticity of claims that are made in the report. This is the most straightforward method (Shu; Sliva; Wang; Tang and Liu 2017). This method uses outside data for the detection of authenticity of the network model. This is also called fact checking method and it is further divided into three sub-categories as follows.

   i) Expert oriented fact checking method demands intellectual presentation for fact checking, it based on human experts to analyze the data and claims in report as right or wrong and determine the final status of news being fake or real.

   ii) Crowd sourcing oriented fact checking method is a crowd sourcing method which works under the ambit of the collective wisdom of the crowd. This concept revolves around the general public instead of exclusive experts to determine the validity of news as fake or real.

   iii) Computational oriented fact checking method is the final model for the network mode method, it advocates the idea of automatic scalable systems to determine the validity and authenticity of any claim.

b) Linguistic Method is a method whereby fake news is determined from the true news with the help of communication channels and letters. Research has shown that liars and truth speakers have different choices of word selection. Liars use other oriental materials whereas truth is spoken with self-orientation. This natural tendency of humans is used to distinguish between fake news and true news; therefore, these properties found in the content of a message can serve as linguistic clues that can detect deception (Rubin 2017).

c) Naïve Bayes classifiers are derived from Bay's Theorems. These calculate the probability of any certain condition happening in any system as something that has happened because something related to it has already occurred (Saxena 2017). Thus, they calculate the imminent outcome of any result by observing the preceding events. These are the machine language clauses. The technique of Naïve bay classifiers is the swift and accessible technique. The biggest downfall of the naïve bay classifiers is that it determines all the features separated which would be perplexing for this technique and may suffer the ability to determine the news because of lack of coordinated analysis.

d) Support Vector Machine (SVM) can be interchangeable with support vector network, also known as SVN. Support Vector Machine works with a superior learnable algorithm. The programmer trains these algorithms special skills. SVM works after it has been trained a special skill. This method classifies the data and it also maximizes the margin between the available columns of two data columns (Brambrick and Aylien 2018. It is a very accurate mode of analysis, along with being more flexible this method can determine the numbers and handle high dimensional storage spaces (Ray; Srivastava; Dar and Shaikh). These are very efficient in terms of data handling and memory storages and the negative aspect is the difficulty in dealing with large data sets because the SVM model requires a lot of time and effort to train algorithm.

e) Semantic Analysis checks the probability of authenticity of the news by comparing the degree of compatibility between personal experience and profile of the content derived from collection of data (Conroy; Rubin and Chen 2015). Semantic analysis works along with Natural Language Processing NLP. This method determines the real and fake news through the process of authorization. This is very essential for language of multiple meanings and close synonyms.

f) Long Short Term Memory Model (LSTM is a model based on visualized network of links and their intermingled relationships determine the authenticity of links (Aldwairi and Alsaadi; 2017). It initiates a thorough search in all the possible outcomes of the search and matches all the information provided to the user, the whole process undergoes a search in which the model determines and identify sites which may contain misleading content such as slang phrases and hyperboles. Such webpages will then be termed as sources of fake news, users are warned against such contents. The rationale behind this model is the idea that the general perspective and fake click baits have longer words than general click baits (Lewis 2011) (this assumption however requires further studies). It determines whether a headline is a potential click bait or normal. Aside from the regular scrutinization of the words and click baits, LSTM also determines punctuation marks and their usage in web pages, associates the bouncing factor of the sites before determining their authenticity. In the case of click baits, they contain information that do not resemble the desired information of the user hence the ratio of bouncing from the sites from the click baits is greater (Chakraborty; Paranjape; Kakarla and Ganguly 2016). The algorithm may present a list of potential fake news to users so they could act accordingly (Aldwairi; Abu-Dalo and Jarrah 2017).

Research in the area of fake news detection through machine learning approaches is advancing at a fast pace. The first fake news challenge stage-1 FNC-1 was organized in June 2017 and the proposal suggested through the challenge has had a success rate of 82%.

**4-WEIGHTED ACCURACY ALGORITHMIC APPROACH**
The weighted accuracy algorithmic approach is a method proposed by this paper as an alternative to the existing methods in fake news detection and reporting.

**4.1** *Method*
This paper uses a combination of seven different machine learning algorithms through Natural Language Processing to analyse the text content of a list of news samples and then predicts whether they are FAKE or REAL based on the weighted accuracies of different algorithms and the maximum prediction accuracy among the models used. Natural Language Processing, NLP is a crucial technique in text analysis which is widely used in fake news detection (Riedel; Augenstein; Spithourakis and Riedel 2017).
PHP language is used for data management while Sklearn is used for training the models in this paper.

### I) Algorithm Training and Best Fit Model Selection

a) Let's import seven algorithms namely,
   i)    Logistic Regression (LR)
   ii)   Linear Discriminant Analysis (LDA)
   iii)  KNeighbors Classifier (KN)
   iv)   Decision Tree Classifier (CART)
   v)    GaussianNB (NB)
   vi)   Support Vector Machine (SVM or SVC)
   vii)  Passive Aggressive Classifier (PAC)

b) Set an algorithm control mechanism to select the models that are suitable for fitting and transforming the dataset (meaning, only algorithms that are able to fit and transform the dataset under study will be used for further analysis).

c) Import and clean up the dataset (such as dropping empty rows and mismatch columns)

d) Set two URL inputs namely, Authentic news source and Unreliable news source

e) Copy and paste the URL of a news source that is generally accepted as authentic and do the same for Unreliable news source into the two URL inputs above respectively.

f) Build a recursive crawler that goes into a website and extract the links and further extract sub-links and put them into a list.

g) The crawler then goes into each link and scrapes the page, looks for news titles and main texts, then saves the extracted content into the existing dataset. Do this step repeatedly (for the same website and other websites) to increase the sample size of the dataset.

h) The extracted data is saved in this format;
    i) Number of words; this is the number of words in the main text of each news article
    ii) Title; the heading or title of each news article
    iii) Text; this is main text content of the news article
    iv) Label; each news from Authentic sources is labeled as REAL whereas those from Unreliable sources are labeled as FAKE

i) Import all seven machine learning algorithms and assign them labels using their abbreviations, e.g. Logistic Regression will be LR
(The algorithm control mechanism will select which algorithms can fit and transform the dataset under study and use them for further analysis)

j) Split the dataset into training and testing samples (usually, 80% and 20% Respectively)

k) Train each of the selected algorithms with the training data and test them separately

l) Record the accuracy, precision, recall, f1-score, support and confusion matrix for each algorithm

m) Find the mean, median, minimum, maximum of the list of accuracies

n) Choose the algorithm with the maximum accuracy as the best fit model for the dataset

## II) -Media Outlet Detection

a) Copy and paste or type media outlet's website address or domain into URL input (stated above) and hit submit or enter

b) The crawler will go into this website and scrape its links recursively and put them into a list

c) The crawler will then scrape all the links and sub-links on the list and then look for news content

d) The news content will be put into a dataframe similar to the one stated above and/or saved to a csv file

e) The best fit model will predict each extracted news content in the dataframe and record the prediction as either FAKE or REAL in the column 'Label'. The content in the column 'Label' will then be transformed from string to integers; thus FAKE=0, REAL=1.

f) Sum all the content in the Label column and divide by the number of elements in the Label column if there is at least one element in the Label column. Thus,
Let authenticity score be S, number of elements in the Label column be n(L) and each label be L, if n(L) ≠ 0, then

$$S = \frac{\sum_{i=0}^{n(L)} L_i}{n(L)} \qquad (1)$$

g) Authenticity score of 1 or 100% is the highest while 0 is the lowest. The higher the authenticity score the closer the cumulative content of the website is to being REAL and vice versa.

h) Set performance constraints for the authenticity score to determine whether the media outlet's website is Authentic or Unreliable. (I used simple percentages as the constraints for simplicity sake).
Let the news dataframe matrix be N, elements of Label column for fake news be Lf and Label column for real news be Lr
There are four possible outcomes;
If n(L) ≠ 0, if $S \geq 75\%$, N(Lf$_i$) = 1, N(Lr$_i$) = 1. Mark all news content on website as Authentic
OR,
If n(L)≠0, if $S \leq 25\%$, N(Lf$_i$) = 0, N(Lr$_i$) = 0. Mark all news content on website as Unreliable
OR,
If n(L)≠0, if $25\% \leq S \leq 75\%$, N(Lf$_i$) = 0, N(Lr$_i$) = 1. Mark each news content on website as Authentic or Unreliable according to the model's prediction
OR
If n(L)=0, the Label column is an empty set. Mark as ∅.

(I used 25% and 75% to get an interquartile range effect of the percentage values of S)

i) From step h) above we can automatically use algorithms to estimate the authenticity of a news outlet's website without having prior knowledge of the website whether being FAKE, REAL or partly FAKE and partly REAL.

j) If the algorithm predicts that the content on the website is partly FAKE and partly REAL, the percentage of fake and real news can be calculated as follows;
Let P(f) be percentage of fake news on website and n(f) be the number of fake news elements in the set L. Then

$$P(f) = \frac{\sum_{i=0}^{n(f)} N(Lf_i)}{n(L)} \qquad (2)$$

Likewise,
Let P(r) be percentage real news on website and n(r) be the number of real news elements in the set L. Then

$$P(r) = \frac{\sum_{i=0}^{n(r)} N(Lr_i)}{n(L)} \qquad (3)$$

k) Let the unacceptable accuracy threshold of best fit model be U(acc), acceptable accuracy threshold of the best fit model be A(acc), maximum of model accuracies of all algorithms be M(acc), mean of accuracies of all algorithms be µ(acc), minimum acceptable mean of accuracies of all algorithms be α(acc). The content on the media outlet will be added to the training dataset to further train the algorithms
if µ(acc) ≥ α(acc);

$$\text{if } M(acc) \geq A(acc)$$

Or if;

$$U(acc) < M(acc) < A(acc) \text{ and } \tfrac{1}{4}A(acc) \leq S \leq \tfrac{3}{4}A(acc)$$

(S is taken from equation (1))

(I used ¼ and ¾ to get an interquartile effect of the percentage values under A(acc))

The constraints are set in order to maximize the quality of the training data thereby increasing the prediction accuracy of the algorithms in subsequent analysis.

### III) Single Link Detection

a) Copy and paste news link into URL input and submit

b) The crawler will go into the page and scrape the news content (title and main text content) and put them into a dataframe and/or save to csv file.

c) The best fit model will predict whether the news content is FAKE or REAL.

d) If μ(acc) ≥ α(acc) and M(acc) ≥ A(acc), The news content together with its label predicted by the best fit model will be added to the training dataset to further train the algorithms.

**4.2 *Results***

In this experiment, I used bbc.com as a source for authentic news and huzlers.com as a source of unreliable news (just for experimental purposes) and cnn.com as a test media outlet website. The initial dataset contained 6335 news contents labeled as FAKE or REAL as shown in figure 1.

```
Unnamed: 0                                             title  \
6330        4490  State Department says it can't find emails fro...
6331        8062  The 'P' in PBS Should Stand for 'Plutocratic' ...
6332        8622  Anti-Trump Protesters Are Tools of the Oligarc...
6333        4021  In Ethiopia, Obama seeks progress on peace, se...
6334        4330  Jeb Bush Is Suddenly Attacking Trump. Here's W...

                                                   text label
6330  The State Department told the Republican Natio...  REAL
6331  The 'P' in PBS Should Stand for 'Plutocratic' ...  FAKE
6332   Anti-Trump Protesters Are Tools of the Oligar...  FAKE
6333  ADDIS ABABA, Ethiopia –President Obama convene...  REAL
6334  Jeb Bush Is Suddenly Attacking Trump. Here's W...  REAL
```

*Figure (1)*

Out of the seven algorithms mentioned above only four of them were able to fit and transform the news text content namely, Linear Regression, KNeighbors Classifiers, Decision Tree Classifier and Passive Aggressive Classifier. The other three could not fit and transform the news content in the dataset and were therefore eliminated from the analysis by the algorithm control mechanism. All four models learn from the dataset and their prediction accuracies according the train and test data are as shown in figure 2 (for the sake of simplicity this paper only considers accuracy scores for further analysis)

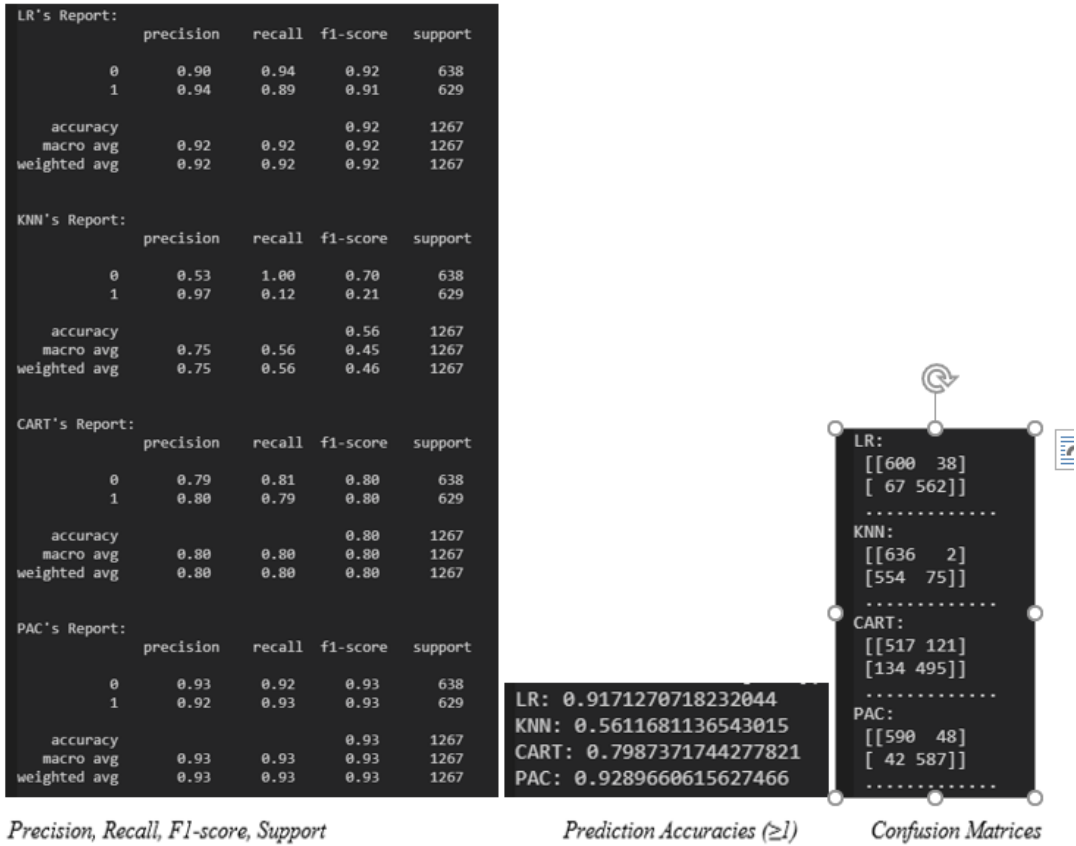

*Figure (2)*

We can see from figure 2 that PAC is the best fit model to qualify for further analysis as it got an accuracy score of ≈0.929, i.e. ≈ 92.9%. KNN got the lowest accuracy score of ≈56.12%
The mean of all model accuracies is 0.801499≈80.15% and the median is ≈85.79%.

The crawler goes into bbc.com, scrapes 562 links and sub-links and extracts 2357 news contents, goes into huzlers.com and scrapes 193 links and sub-links and extracts only 24 news contents (the number of links, sub-links and news content extracted will be based on the quality of their contents as determined by the crawler). Both extracted news contents are added to the dataset, so the are 2357+24 = 2381 news contents for the algorithms to learn from.

The crawler then goes into cnn.com, scrapes 528 links and sub-links and extracts 23 news contents. All four model predict whether the 23 contents are FAKE or REAL news.
The results are as shown in figure 3.

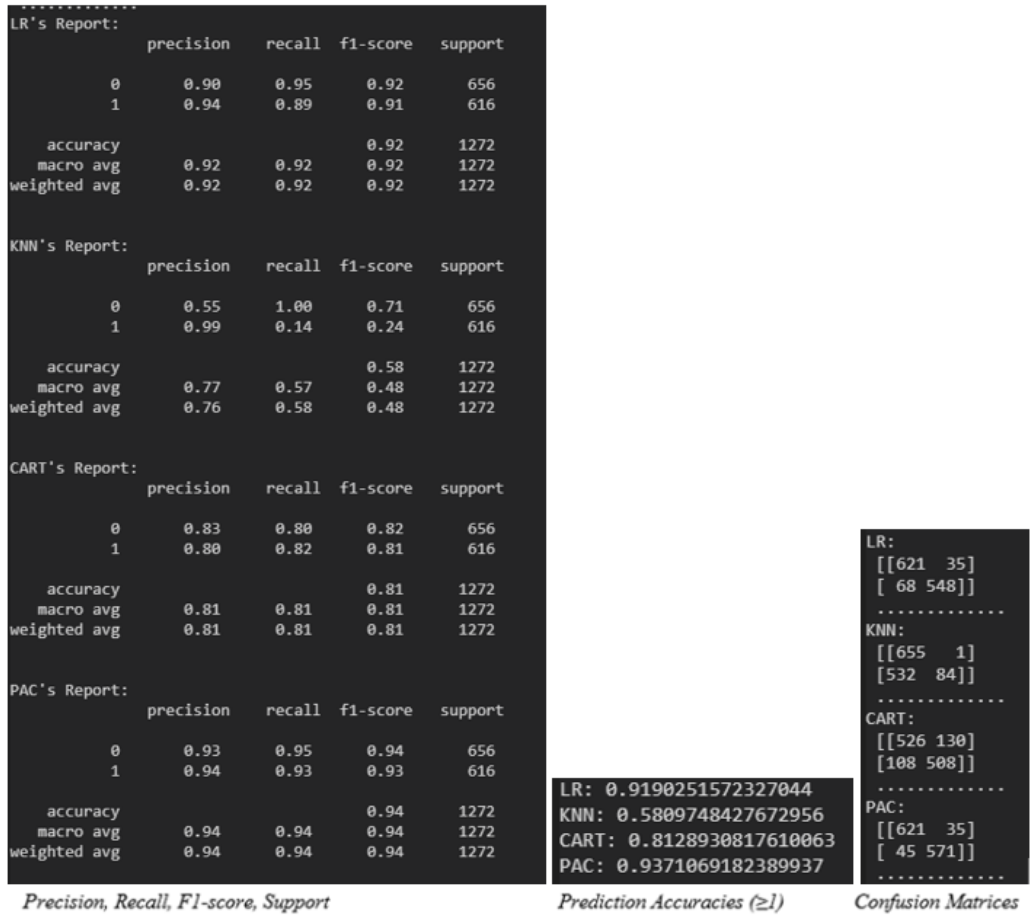

*Figure (3)*

We can see clearly from figure 3 that the accuracy scores have improved from the previous figures. PAC is still the best fit model, however, the other models also performed better than their previous scores. PAC got an accuracy score of ≈0.937, i.e. ≈ 93.7%. KNN still got the lowest accuracy score of ≈58.09% (better than previous 56.12%).
The mean of all model accuracies is 0.8125≈81.25% and the median is ≈86.59%.

This shows the combined algorithm performance used as the indicator to determine which news is FAKE or REAL and which data is qualified to be added to further train the models for further analysis has shown to increase the efficiency of all the algorithms used (further study is required to determine how the constraints α(acc) and A(acc) can be efficiently deduced).

I then proceed to do another out of sample test using a single news link to check whether it is FAKE or REAL news. I copy the link https://news.yahoo.com/spains-catalonia-stricter-measures-coronavirus-080017814.html from news.yahoo.com and paste it into the URL input and submit. The results is as shown in figure 4.

```
PAC, with accuracy rate of 0.9387274155538099 is best fit model for the analysis
 Best fit model: PAC
 Accuracy: 93.87274155538098%
Prediction of Stop partying or we may go back into lockdown, regional chief tells young Catalans:
Model predicts the news is REAL
Prediction rate: High
Stop partying or we may go back into lockdown, regional chief tells young Catalans
is possibly REAL news
Saved..

                                           text  label
6358  WeatherTrack the latest weather stories and sh...   FAKE
6359  The latest on President Donald Trump, the Whit...   FAKE
6360                    The latest on the Supreme Court.   REAL
6361  CNN holds elected officials and candidates acc...   FAKE
6362  BARCELONA (Reuters) - Young Catalans should st...   REAL
```

*Figure (4)*

We can see from figure 4 that PAC is still the best fit model for the analysis. It predicted that the news is REAL and the constraints mechanism described in the previous chapter also put the news as REAL so the news content is deemed REAL. The news content is qualified to be saved for further training of the algorithms based on constraints set by the mechanism of weighted accuracies of all four models and the maximum accuracy of the four accuracies, as described in the previous chapter.

The same test is done using the link https://www.infowars.com/bill-gates-vaccines-transhumanism-dark-secrets-you-need-to-know/ and the results is as shown in figure 5.

```
Prediction of Bill Gates, Vaccines & Transhumanism: Dark Secrets You Need To KnowWatch LiveFeaturedRel
ated ArticlesSearchSearchSide Hero AdToday on the ShowListen to the Alex Jones Show - Here's HowWatch
 the ShowWatch Live NowTop StoriesLatest StoriesSignupGet InformedFrom the storeVisit Our StoreWatch t
he newsILLUSTRATIONIllustrationPollPOLLS:
Model predicts the news is FAKE
Prediction rate: High
Bill Gates, Vaccines & Transhumanism: Dark Secrets You Need To KnowWatch LiveFeaturedRelated ArticlesS
earchSearchSide Hero AdToday on the ShowListen to the Alex Jones Show - Here's HowWatch the ShowWatch
 Live NowTop StoriesLatest StoriesSignupGet InformedFrom the storeVisit Our StoreWatch the newsILLUSTR
ATIONIllustrationPollPOLLS
is possibly FAKE news
Saved..

                                           text  label
6359  The latest on President Donald Trump, the Whit...   FAKE
6360                    The latest on the Supreme Court.   REAL
6361  CNN holds elected officials and candidates acc...   FAKE
6362  BARCELONA (Reuters) - Young Catalans should st...   REAL
6363  David Knight is joined by Dr. Sherri Tenpenny ...   FAKE
```

*Figure (5)*

PAC predicts the news is FAKE, the weighted effect of the four algorithms allows the content to be saved, the prediction accuracy is marked as high, and not to sound prejudicial, this news source is listed on Wikipedia as a fake news source (Wiki 2020).

**4.3** *Discussion*
The weighted accuracy algorithmic approach in fake news detection is different from the existing mechanisms in the sense that it automates the algorithm selection process and at the same time takes into account the performance of all used models including the less performing algorithms. This helps to reduce overfitting by a singular algorithm that seems to be performing greatly. Also, this paper does not claim that the mechanism used here is the perfect solution for fake news detection; it is nevertheless an alternative to specialized algorithm detection mechanisms such as LSTM etc. More research is needed to show how to determine the constraints α(acc) and A(acc) more precisely.

**5-CONCLUSION**

News is a critical factor for decision making. In this digital era, not only has misinformation become a marketing technique or public manipulative tool but has also been used by extremists which could bring about serious calamities such as public unrest if not handled with care. Therefore, it is imperative to counter any effect on the internet or in the area of information technology that may promote fake or false data and spread confusion in the society. This paper proposes a simple and effective tool based on weighted accuracy machine learning algorithms to determine potentially false or misleading information on the internet. The mechanism used in this paper has shown to improve the efficiency of detection through the combined performance of different algorithms thereby correcting overfitting that may produce overconfident results if singular or specialized algorithms are used.
A call for a more hygienic information environment is crucial to the socio-economic and geopolitical system of the global economy and internationalism as a whole.

**Abbreviations**

AI: Artificial Intelligence

U.S: United States

EU: European Union

PHP - Hypertext Preprocessor is a widely-used, open source scripting language

Sklearn - Scikit-learn is a free machine learning library which features algorithms

**Figure legends**: Not applicable

**Figures**: 5

**External Dataset**: Not Applicable

## 6-REFERENCES

1. Fake News Wikipedia, As of May 25 2020 https: //en.wikipedia.org/wiki/Fakenews

2. Chen, Y., Conroy, N. J., & Rubin, V. L. (2015). News in an Online World: The Need for an "Automatic Crap Detector". In the Proceedings of the Association for Information Science and Technology Annual Meeting (ASIST2015).

3. Feng and Hirst, 2013; Markowitz and Hancock, 2014; Ruchansky,2017

4. Bachenko, 2008; Larcker and Zakolyukina,2012). 6. Rubin, V.L., Chen, Y., Conroy, N.J., 2015. Deception detection for news: Three types of fakes, in: Proceedings of the 78th ASIS&T Annual Meeting: Information Science with Impact: Research in and for the Community,

5. American Society for Information Science, Silver Springs, MD, USA. pp. 83:1–83:4. November 2015, URL: http://dl.acm.org/citation.cfm?id=2857070.2857153.

6. Balmas, M., 2014. When fake news becomes real: Combined exposure to multiple news sources and political attitudes of inefficacy, alienation, and cynicism. Communication Research 41, 430–454

7. ICFNDS, Malware detection using DNS records and domain name features, June 2018 https://doi.org/10.1145/3231053.3231082.

8. Nah, F.F.H., 2015. Fake-website detection tools: Identifying elements that promote individuals use and enhance their performance 1. introduction.

9. Pogue, D., 2017. How to stamp out fake news. Scientific American 316, 24–24.

10. Hassid, J., 2011. Four models of the fourth estate: A typology of contemporary Chinese journalists. The China Quarterly 208.

11. Conroy, N., Rubin, V., & Chen, Y. (2015). Automatic deception detection: Methods for finding fake news. Proceedings of the Association for Information Science and Technology, 52(1), 1-4

12. Shu, K., Sliva, A., Wang, S., Tang, J., & Liu, H. (2017). Fake News Detection on Social Media: A Data Mining Perspective. ACM SIGKDD Explorations Newsletter, 19(1), 22-36.

13. Rubin, V. (2017). Deception detection and rumor debunking for social media. Handbook of Social Media Research Methods.

14. Saxena, R. (2017). How the Naive Bayes Classifier works in Machine Learning. Retrieved October 20, 2017, from https://dataaspirant.com/2017/02/06/naive-bayes-classifier-machine-learning/

15. Brambrick, Aylien, N. (n.d.). KDnuggets. Retrieved February 20, 2018, from https://www.kdnuggets.com/2016/07/support-vector-machines-simple-explanation.html

16. Ray, S., Srivastava, T., Dar, P., & Shaikh, F. (2017). Understanding Support Vector Machine algorithm from examples (along with code). Retrieved March 2, 2018, from https://www.analyticsvidhya.com/blog/2017/09/understaing-support-vector-machine-example-code/

17. Conroy, N., Rubin, V., & Chen, Y. (2015). Automatic deception detection: Methods for finding fake news. Proceedings of the Association for Information Science and Technology, 52(1), 1-4.

18. Aldwairi, M., Alsaadi, H.H., 2017.Flukes: Autonomous log forensics, intelligence and visualization tool, in: Proceedings of the International Conference on Future Networks and Distributed Systems, ACM, New York, NY, USA. pp. 33: 133: 6. URL: http://doi.acm.org/10.1145/3102304.3102337,doi:10.1145/3102304.3102337.

19. Lewis, S., 2011. Journalists, social media, and the use of humor on twitter. The Electronic Journal of Communication / La Revue Electronic de Communication 21, 1–2.

20. Chakraborty, A., Paranjape, B., Kakarla, S., Ganguly, N., 2016. Stop clickbait: Detecting and preventing clickbaits in online news media, in: 2016 IEEE/ACM International Conference on Advances in Social Networks Analysis and Mining (ASONAM), pp. 9–16. doi:10.1109/ASONAM.2016.7752207.

21. Aldwairi, M., Abu-Dalo, A.M., Jarrah, M., 2017a.Patternmatchingofsignature-basedidsusingmyersalgorithmundermapreduceframe-work. EURASIPJ.InformationSecurity2017, 9.URL: http://dblp.uni-trier.de/db/journals/ejisec/ejisec2017.html#AldwairiAJ17.

22. Riedel, B., Augenstein, I., Spithourakis, G.P., Riedel, S., 2017. A simple but tough-to-beat baseline for the fake news challenge stance detection task. CoRR abs/1707.03264. URL: http://arxiv.org/abs/1707.03264, arXiv:1707.03264.

23) Wikipedia, as of July 28, 2020 https://en.wikipedia.org/wiki/InfoWars